\documentclass[11pt,a4paper]{article}
\usepackage[hyperref]{naaclhlt2019}
\usepackage{times}
\usepackage{latexsym}
\usepackage{booktabs}
\usepackage{xspace}
\usepackage{hyperref}
\usepackage{graphicx}
\usepackage{url}
\usepackage{paralist}
\usepackage{multirow}
\usepackage{relsize}
\usepackage[normalem]{ulem}
\usepackage[T1]{fontenc}
\usepackage{amsmath}
\usepackage{amsthm}
\usepackage{expex}
\aclfinalcopy 

\newcommand{\class}[1]{\textsc{#1}\xspace}
\newcommand{\sa}[1]{\textit{#1}\xspace}
\newcommand{\policy}[1]{\textbf{#1}\xspace}

\newcommand\tabref[2][]{Table#1~\ref{#2}}
\newcommand\figref[1]{Figure~\ref{#1}}
\newcommand\secref[1]{Section~\ref{#1}}

\newcommand{\uttseg}{\textbf{\big{|}}\xspace}

\title{Target Based Speech Act Classification in Political Campaign Text}

\author{Shivashankar Subramanian \qquad Trevor Cohn \qquad Timothy
  Baldwin \\School of Computing and Information Systems\\ The University
  of Melbourne \\
 {\small \url{shivashankar@student.unimelb.edu.au} \qquad \url{{t.cohn, tbaldwin}@unimelb.edu.au}}}
\date{}

\begin{document}
\maketitle
\begin{abstract}
  We study pragmatics in political campaign text, through analysis of
  speech acts and the target of each utterance. We propose a new annotation schema incorporating domain-specific speech acts, such as \sa{commissive-action}, and present a novel annotated corpus of media
  releases and speech transcripts from the 2016 Australian election
  cycle. We show how speech acts and target referents can be modeled as sequential classification, and evaluate several techniques, exploiting contextualized word representations, semi-supervised learning, task dependencies and speaker meta-data.
\end{abstract}

\section{Introduction}
\label{intro}
Election campaign text is a core artifact in political analysis. Campaign
communication can influence a party's reputation, credibility, and competence, which are primary factors in voter decision making \cite{fernandez2014and}. 
Also, modeling the discourse is key to measuring the role of party in \textit{constructive democracy} --- to engage in constructive discussion with other parties in a democracy \cite{APSA2017}. 

Speech act theory \cite{Austin, Searle} can be used to study such pragmatics in political campaign text.  Traditional speech act classes have been studied to analyze how people engage with elected members \cite{CSCW2014}, and how elected members engage in discussions \cite{DeepBlue2018}, with a particular focus on pledges \cite{Artes, Naurin2011, Naurin2014, APSA2017}. Also, election manifestos have been analyzed for prospective and retrospective messages \cite{MCC2018}. In this work, we combine traditional speech acts with those proposed by political scientists to study political discourse, such as \textit{specific pledges}, which can also help to verify the pledges' fulfilment after an election \cite{thomson2010program}. 

In addition to speech acts, it is important to identify the target of each utterance --- that is, the political party referred to in the text --- in order to determine the discourse structure. Here, we study the effect of jointly modeling the speech act and target referent of each utterance, in order to exploit the task dependencies. That is, this paper is an application of discourse analysis to the pragmatics-rich domain of political science, to determine the intent of every utterance made by politicians, and in part, automatically extract pledges at varying levels of specificity from campaign speeches and press releases.

We assume that each utterance is associated with a unique speech act
(similar to \citet{zhao2017joint}) and target party,\footnote{\citet{zhao2017joint} do not address the target referent classification task in their work.} meaning that a
sentence with multiple speech acts and/or targets must be segmented into
component utterances. Take the following example, from the Labor Party:
\ex[aboveexskip=0.9em,belowexskip=0.9em]<overallex>
 {Labor will contribute \$43 million towards the Roe Highway project \uline{and we call on the WA Government to contribute funds to get the project underway.}}
\xe
The example is made up of two utterances (with and without an
underline), belonging to speech act types
\sa{commissive-action-specific} and \sa{directive}, referring to
different parties (\class{Labor} and \class{Liberal}), resp.  In our
initial experiments, we perform target based speech act classification
(i.e.\ joint speech act classification and determination of the target
of the utterance) over gold-standard utterance data (\secref{eval}), but
return to perform automatic utterance segmentation along with target
based speech act classification (\secref{Segment}).

\begin{table*}[!htb]
\centering
\begin{smaller}
\begin{tabular}{p{7.9cm}lll}
\toprule
 Utterance & Speech act & Target party & Speaker\\
\midrule
Tourism directly and indirectly supports around 38000 jobs in TAS.
& \sa{assertive} &  \class{None} &  \class{Labor} \\ \\[-0.8em]
We will invest \$25.4 million to increase forensics and intelligence assets for the Australian Federal Police & \sa{commissive-action-specific} & \class{Liberal} & \class{Liberal} \\ \\[-0.8em]
Labor will prioritise the Metro West project if elected to government. & \sa{commissive-action-vague} &  \class{Labor} & \class{Labor} \\ \\[-0.8em]
A Shorten Labor Government will create 2000 jobs in Adelaide.
& \sa{commissive-outcome} & \class{Labor} & \class{Labor} \\ \\[-0.8em]
Federal Labor today calls on the State Government to commit the final \$75 million to make this project happen. &  \sa{directive} & \class{Liberal} & \class{Labor} \\ \\[-0.8em]
Good morning everybody. & \sa{expressive} & \class{None} & \class{Labor}\\ \\[-0.8em]
The Coalition has already delivered a \$2.5 billion boost to our law enforcement and security agencies.
& \sa{past-action} &  \class{Liberal} & \class{Liberal}\\ \\[-0.8em]
Malcolm Turnbull's health cuts will rip up to \$1.4 billion out of Australians' pockets every year & \sa{verdictive} & \class{Liberal} & \class{Labor}\\ 
\bottomrule
\end{tabular}
\end{smaller}
\caption{Examples with speech act and target party classes. ``Speaker''
  denotes the party making the utterance.}
\label{tab:clra}
\end{table*}

While speech act classification has been applied to a wide range of domains, its application to political text is relatively new. Most speech act analyses in the political domain have relied exclusively on manual annotation, and no labeled data has been made available for training classifiers. As it is expensive to obtain large-scale annotations, in addition to developing a novel annotated dataset, we also experiment with a semi-supervised approach by utilizing unlabeled text, which is easy to obtain.

The contributions of this paper are as follows: (1) we introduce the
novel task of target based speech act classification to the analysis of
political discourse; (2) we develop and release a dataset (can be found here \url{https://github.com/shivashankarrs/Speech-Acts}) based on
political speeches and press releases, from the two major parties ---
Labor and Liberal --- in the 2016 Australian federal election cycle; and (3) we
propose a semi-supervised learning approach to the problem by augmenting the training data with in-domain unlabeled text.

\section{Related Work}
The recent adoption of NLP methods has led to significant advances in the field of computational social science \cite{lazer2009life}, including political science \cite{grimmer2013text}. With the increasing availability of datasets and computational resources, large-scale comparative political text analysis has gained the attention of political scientists \cite{lucas2015computer}. One task of particular importance is the analysis of the functional intent of utterances in political text. Though it has received notable attention from many political scientists (see \secref{intro}), the primary focus of almost all work has been to derive insights from manual annotations, and not to study computational approaches to automate the task.

Another related task  in the political communication domain is reputation defense, in terms of party credibility. Recently, \citet{duthie2018deep} proposed an approach to mine ethos support/attack statements from UK parliamentary debates, while \citet{W18-5214} focused on classifying sentences from Question Time in the Canadian parliament as defensive or not. In this work, our source data is speeches and press releases in the lead-up to a federal election, where we expect there to be rich discourse and interplay between political parties.

Speech act theory is fundamental to study such discourse and pragmatics \cite{Austin, Searle}. A speech act is an illocutionary act of conversation and reflects shallow discourse structures of language. Due to its predominantly small-data setting, speech act classification approaches have generally relied on bag-of-words models \cite{qadir2011classifying, vosoughi2016tweet}, although recent approaches have used deep-learning models through data augmentation \cite{joty2016speech} and learning word representations for the target domain \cite{joty2018speech}, outperforming traditional bag-of-words approaches. 

Another technique that has been applied to compensate for the sparsity of labeled data is semi-supervised learning, making use of auxiliary unlabeled data, as done previously for speech act classification in e-mail and forum text \cite{jeong2009semi}. \citet{zhang2012towards} also used semi-supervised methods for speech act classification over Twitter data. They used transductive SVM and graph-based label propagation approaches to annotate unlabeled data using a small seed training set. \citet{joty2018speech} leveraged out-of-domain labeled data based on a domain adversarial learning approach. In this work, we focus on target based speech act analysis (with a custom class-set) for political campaign text and use a deep-learning approach by incorporating contextualized word representations \cite{peters2018deep} and a cross-view training framework \cite{clark2018semi} to leverage in-domain unlabeled text.

\section{Problem Statement}
\label{PS}

Target based speech act classification requires the segmentation of
sentences into utterances, and labelling of those utterances according
to speech act and target party. In this work we focus primarily on
speech act and target party classification.

Our speech act coding schema is comprised of: \sa{assertive},
\sa{commissive}, \sa{directive}, \sa{expressive}, \sa{past-action}, and
\sa{verdictive}. An \sa{assertive} commits the speaker to something
being the case. With a \sa{commissive}, the speaker commits to a future
course of action. Following the work of \citet{Artes} and \citet{Naurin2011}, we
distinguish between \sa{action} and \sa{outcome} commissives. Action
commissives (\sa{commissive-action}) are those in which an action is to
be taken, while outcome commissives (\sa{commissive-outcome}) can be
defined as a description of reality or goals. Secondly, similar to
\citet{Naurin2014} we also classify action commissives into vague
(\sa{commissive-action-vague}) and specific
(\sa{commissive-action-specific}), according to their specificity. This
distinction is also related to text specificity analysis work addressed
in the news \cite{louis2011automatic} and classroom discussion
\cite{lugini2017predicting} domains. A \sa{directive} occurs when the
speaker expects the listener to take action in response. In an
\sa{expressive}, the speaker expresses their psychological state,
while a \sa{past-action} denotes a retrospective action of the target
party, and a \sa{verdictive} refers to an assessment on prospective or
retrospective actions.

Examples of the eight speech act classes
are given in \tabref{tab:clra}, along with the target
party (\class{Labor}, \class{Liberal}, or \class{None}), indicating
which party the speech act is directed towards, and the ``speaker'' party
making the utterance (information which is provided for every utterance).

\subsection{Utterance Segmentation}
\label{Segmentex}
Sentences are segmented both in the context of speech act and target
party --- when a sentence has utterances belonging to more than one
speech act or/and more than one target. For example, the following sentence conveys a pledge (\sa{commissive-outcome}) followed by the party's belief (\sa{assertive}), with the utterance boundary indicated by \uttseg: 
\ex[aboveexskip=0.75em,belowexskip=0.75em]<seg1>
{We will save Medicare \uttseg because Medicare is more than just a standard of health.} 
\xe
Further, the following (from the Labor party) has segments comparing
\class{Labor} and \class{Liberal}:
\ex[aboveexskip=0.75em,belowexskip=0.75em]<seg2>
{Our party is united -- \uttseg the Liberals are not united.}  
\xe

\section{Election Campaign Dataset}
\label{dataset}

We collected media releases and speeches from the two major
Australia political parties --- Labor and Liberal --- from the 2016
Australian federal election campaign. A statistical breakdown of the dataset is given in \tabref{tab:ds}. We compute agreement over 15
documents, annotated by two independent annotators, with disagreements
resolved by a third annotator. The remaining documents are annotated
by the two main annotators without redundancy. Agreement between the two annotators for utterance segmentation based on exact boundary match using Krippendorff's alpha ($\alpha$) \cite{krippendorff2011computing} is 0.84. Agreement statistics for the classification tasks
\cite{cohen1960coefficient, carletta1996assessing} are given in
\tabref[s]{tab:sa} and \ref{tab:tp}.

 \begin{table}[!t]
  \centering
  \begin{tabular}{ c c c c}
  \toprule
  \# Doc & \# Sent & \# Utt & Avg Utterance Length \\
    \midrule
    258 & 6609 & 7641 & 19.3 \\
    \bottomrule
  \end{tabular}
  \caption{Dataset Statistics: number of documents, number of sentences, number of utterances, and average utterance length}
 \label{tab:ds}
\end{table}

 \begin{table}[!t]
  \centering
  \begin{small}
  \begin{tabular}{ l r c }
  \toprule
    Speech act & \% & Kappa ($\kappa$)\\
    \midrule
    \sa{assertive}    & 40.8 & 0.85\\
    \sa{commissive-action-specific}  &  12.4 & 0.84 \\
    \sa{commissive-action-vague} &  6.6 & 0.73 \\
    \sa{commissive-outcome} &  4.9 &  0.72 \\    
    \sa{directive}  &  1.7 & 0.92 \\    
    \sa{expressive}  &  1.9 & 0.88 \\    
    \sa{past-action}  &  6.3 & 0.76 \\    
    \sa{verdictive}  &  25.4 & 0.82 \\    
    \bottomrule
  \end{tabular}
  \end{small}
  \caption{Speech act agreement statistics}
 \label{tab:sa}
\end{table}

 \begin{table}[!t]
  \centering
  \begin{small}
  \begin{tabular}{ lrc }
  \toprule
    Target party & \% & Kappa ($\kappa$)\\
    \midrule
   \class{Labor}  &  45.9 & 0.92 \\    
    \class{Liberal}  &  39.1 & 0.90 \\    
     \class{None}  &  15.0 & 0.86 \\    
    \bottomrule
  \end{tabular}
  \end{small}
  \caption{Target party agreement statistics}
 \label{tab:tp}
\end{table}

\begin{table*}[!t]
  \centering
  \small
  \begin{tabular}{l l c c c cc}
\toprule
\multirow{2}{*}{ID} & \multirow{2}{*}{Approach} & \multicolumn{2}{c}{Speech act} && %
    \multicolumn{2}{c}{Target party}\\ \\[-0.95em]
\cline{3-4} \cline{6-7} \\[-0.95em]
  & &  Accuracy & Macro-F1 && Accuracy & Macro-F1\\
\midrule  
1 & Meta$_{\textbf{\smaller{naive}}}$ &--- & --- &&\,\,\, 0.55 & 0.43 \\ \\[-0.95em]
2 & SVM$_{\textbf{\smaller{BoW}}}$  & 0.56 &  0.41 &&\,\,\,\,\,\,\,\,\, 0.60*$^{1}$ & \,\,\,\,\,\,0.56*$^{1}$ \\  \\[-0.95em]
3 & MLP$_{\textbf{\smaller{BoW}}}$  & \,\,\,\,\,0.60*$^{2}$  &  \,\,\,\,\,0.47*$^{2}$  &&\,\,\,\,\,\,\,\,\, 0.61*$^{1}$ & \,\,\,\,\,\,0.57*$^{1}$ \\  \\[-0.95em]
4 & DAN$_{\textbf{\smaller{GloVe}}}$  & 0.53 & 0.30 &&\,\,\, 0.59 & 0.54 \\  \\[-0.95em]
5 & GRU$_{\textbf{\smaller{GloVe}}}$ & 0.56  & 0.46  &&\,\,\, 0.58 & 0.55 \\  \\[-0.95em]
6 & biGRU$_{\textbf{\smaller{GloVe}}}$ & 0.57 & 0.48 &&\,\,\, 0.59 & 0.56 \\   \\[-0.95em]
7 & MLP$_{\textbf{\smaller{ELMo}}}$& \,\,\,\,\,\,0.62*$^{3}$ & \,\,\,\,\,\,0.53*$^{3}$ &&\,\,\, 0.58 & 0.57 \\ \\[-0.95em]
8 & biGRU$_{\textbf{\smaller{ELMo}}}$ & \,\,\,\,\,\,\textbf{0.68}*$^{7}$ & \,\,\,\,\,\,\textbf{0.57}*$^{7}$ &&\,\,\, \,\,\,\,\,\,\,\,\,\,\,\,\,0.63*$^{2,3,7}$ & \,\,\,\,\,\,\,\,\,\,\,\,\,0.60*$^{2,3,7}$ \\  \\[-0.95em]
\midrule \\[-0.9em]
9 & biGRU$_{\textbf{\smaller{ELMo + CVT$_\text{fwd}$}}}$ & 0.66 & 0.55 &&\,\,\, 0.63 & 0.58 \\  \\[-0.95em]
10 & biGRU$_{\textbf{\smaller{ELMo + CVT$_\text{fwdbwd}$}}}$ & \textbf{0.68} & 0.54 &&\,\,\, 0.61 & 0.56 \\  \\[-0.95em]
11 & biGRU$_{\textbf{\smaller{ELMo + CVT$_\text{worddrop}$}}}$ & \textbf{0.69} & \textbf{0.57} &&\,\,\,\,\,\,\,\, 0.66*$^{8}$ & 0.60 \\  \\[-0.95em]
\midrule \\[-0.9em]
12 & biGRU$_{\textbf{\smaller{ELMo + CVT$_\text{worddrop}$ + Multi}}}$ & \textbf{0.69} & \textbf{0.58} &&\,\,\, 0.65 & 0.60 \\ \\[-0.95em]
\midrule \\[-0.9em]
13 & biGRU$_{\textbf{\smaller{ELMo + CVT$_\text{worddrop}$ + Meta}}}$  & \textbf{0.68}& \textbf{0.58} &&\,\,\,\,\,\,\,\,\,\, \textbf{0.71}*$^{11}$ & \,\,\,\,\,\,\,\,\,\,\,\textbf{0.66}*$^{8, 11}$ \\
 \bottomrule
   \end{tabular}
   \caption{Classification results showing average performance based on 10 runs. * indicates results significantly better than the indicated approaches (based on ID in the table) according to a paired t-test ($p<0.05$). Boldface shows the overall best results and results insignificantly different from the best. Meta$_{\textbf{\smaller{naive}}}$ is not applicable for speech act classification. Note that all approaches use gold-standard segmentation for evaluation.}
  \label{tab:res}
\end{table*}

\section{Proposed Approach}
\label{models}

Our approach to labeling utterances for speech act and
target party classification is as follows. 
Utterances are first represented as a sequence of word embeddings,
and then using a bidirectional Gated Recurrent Unit
(``biGRU'': \citet{cho2014learning}). The representation of each utterance
is set to the concatenation of the last
hidden state of both the forward and backward GRUs, 
\mbox{$\mathbf{h}_{i} = \left[\overrightarrow{\mathbf{h}}_{i}, \overleftarrow{\mathbf{h}}_{i}\right]$}.
After this, the model has a softmax output layer. This network is trained for both the speech act (eight class) and target party (three class) classification tasks, minimizing cross-entropy loss, denoted as $\mathcal{L}_{S}$ and $\mathcal{L}_{T}$ respectively.  \\[-0.95em]

Our approach has the following components:
\paragraph{\textbf{ELMo word embeddings}} (``biGRU$_{\textbf{\smaller{ELMo}}}$''): 
  As word embeddings we use a 1024d
  learned linear combination of the internal states of a bidirectional
  language model \cite{peters2018deep}.
  
 \paragraph{\textbf{Semi-supervised Learning}:} We employ a cross-view training approach \cite{clark2018semi} to leverage a larger volume of unlabeled text. Cross-view training is a kind of teacher--student method, whereby the model ``teaches'' another ``student'' model to classify unlabelled data. The student sees only a limited form of the data, e.g., through application of noise \cite{sajjadi2016regularization, wei2018improving}, or a different view of the input, as used herein. This procedure regularises the learning of the teacher to be more robust, as well as increasing the exposure to unlabeled text. 
 
We augment our dataset with over 36k sentences from Australian Prime Minister candidates' election speeches.\footnote{\url{https://primeministers.moadoph.gov.au/collections/election-speeches}} 
On these unlabeled examples, the model's probability distribution over targets $p_{\theta}(y|s)$ is used to fit auxiliary model(s), $p_{\omega}(y|s)$, by minimising the Kullback-Leibler (KL) divergence, $\text{KL}(p_{\theta}(y|s), p_{\omega}(y|s))$.
This consensus loss component, denoted $\mathcal{L}_{\text{unsup}}$, is added to the supervised training objective ($\mathcal{L}_{S}$
or $\mathcal{L}_{T}$).

We evaluate the following auxiliary models:\footnote{Note that auxliary models share parameters with the corresponding components of main (teacher) model, with the exception of their output layers.}
\begin{compactitem}
\item a forward GRU (``biGRU$_{\textbf{\smaller{CVT$_\text{fwd}$}}}$''); \\[-0.95em]
\item separate forward and backward GRUs (``biGRU$_{\textbf{\smaller{CVT$_\text{fwdbwd}$}}}$''); and \\[-0.95em]
\item a biGRU with word-level dropout (``biGRU$_{\textbf{\smaller{CVT$_\text{worddrop}$}}}$''). \\[-0.95em]
\end{compactitem}
The intuition is that the student models only have access to restricted views of the data on which the teacher network is trained, and therefore
this acts as a regularization factor over the unlabeled data when learning the teacher model.

\paragraph{\textbf{Multi-task Learning}} (``biGRU$_{\textbf{\smaller{Multi}}}$''):
  For speech act classification, target party classification is
  considered as an auxiliary task, and vice versa. Accordingly, a separate
  model is built for each task, with the other task as an auxiliary
  task, in each case using a linearly weighted objective
  $\mathcal{L}_S + \alpha \mathcal{L}_T$, where $\alpha \geq 0$ is tuned
  separately in each application. The intuition here is to capture the dependencies between the tasks, e.g., \sa{commissive} is relevant to the Speaker party only.
 
\paragraph{\textbf{Meta-data}} (biGRU$_{\textbf{\smaller{Meta}}}$): We
  concatenate a binary flag encoding the speaker party
  ($\mathbf{m}_{i}$) alongside the utterance embedding $\mathbf{h}_{i}$, i.e.,
  $\left[\mathbf{h}_{i}, \mathbf{m}_{i}\right]$. This representation is passed through a 
  hidden layer with ReLU-activation, then projected onto a output layer with softmax activation for both the classification tasks.

\section{Evaluation}
\label{eval}

We compare the models presented in \secref{models} with the following baseline approaches:
\begin{itemize}

\item Support Vector Machine (``SVM$_{\textbf{\smaller{BoW}}}$'') with with unigram term-frequency representation.

\item Multi-layer perceptron (``MLP$_{\textbf{\smaller{BoW}}}$'') with unigram term-frequency representation.
  
\item Deep Averaging Networks (``DAN$_{\textbf{\smaller{GloVe}}}$'') \cite{iyyer2015deep}, GRU (``GRU$_{\textbf{\smaller{GloVe}}}$''), and biGRU
  (``biGRU$_{\textbf{\smaller{GloVe}}}$'') with pre-trained 300d
  GloVe embeddings \cite{pennington2014glove}.
 
\item MLP with average-pooling over pre-trained ELMo word embeddings
  (``MLP$_{\textbf{\smaller{ELMo}}}$'').
\item Using speaker party as the predicted target party (``Meta$_{\textbf{\smaller{naive}}}$''). 
\end{itemize}

\begin{table}[!t]
      \centering
      \begin{smaller}
     \begin{tabular}{l c c}
    \toprule
       Speech act & MLP$_\text{ELMo}$ & Our approach \\
       \midrule  
       \sa{assertive}  & 0.77 &  0.80\\
       \sa{commissive-action-specific}  & 0.65  &  0.69 \\
       \sa{commissive-action-vague} & 0.45 &  0.48 \\     
       \sa{commissive-outcome} & 0.28 &  0.39   \\
       \sa{directive}  & 0.58 &  0.59   \\
       \sa{expressive}  & 0.55 & 0.58\\ 
       \sa{past-action}  & 0.45 & 0.48 \\
       \sa{verdictive}  &0.48 &0.61  \\  
     \bottomrule
   \end{tabular}
  \end{smaller}
  \caption{Speech act class-wise F1 score.}
  \label{tab:dres}
\end{table}

\begin{table}[!t]
\centering
      \begin{small}
      \begin{tabular}{l c c}
\toprule
Target party & biGRU$_\text{ELMo}$ & Our approach\\
 
\midrule  
    \class{Labor} & 0.68 & 0.74 \\
    \class{Liberal}& 0.65 & 0.75\\
    \class{None} & 0.46 &0.48\\
     \bottomrule
   \end{tabular}
  \end{small}
  \caption{Target party class-wise F1 score.}
  \label{tab:drest}
\end{table}

We average results across 10 runs with 90\%/10\% training/test random
splits. Hyper-parameters are tuned over a 10\% validation set randomly
sampled and held out from the training set. We evaluate using accuracy and
macro-averaged F-score, to account for class-imbalance. We compare the baseline approaches against our proposed approach (different components given in \secref{models}). We evaluate the effect of each component by adding them to the base model (biGRU$_{\textbf{\smaller{ELMo}}}$), e.g., biGRU model with ELMo embeddings and word-level dropout based semi-supervised approach is given as biGRU$_{\textbf{\smaller{ELMo + CVT$_\text{worddrop}$}}}$.  Results for speech act and target party
classification are given in \tabref{tab:res}. The corresponding
class-wise performance for both speech act and target party tasks with our approach (biGRU$_{\textbf{\smaller{ELMo + CVT$_\text{worddrop}$} + 
Meta}}$) compared against the competitive approach from Table \ref{tab:res} is given in \tabref{tab:dres} and \tabref{tab:drest}
respectively (and also discussed further in \secref{error}). All the approaches are evaluated with the gold-standard segmentation. Utterance
segmentation is discussed in \secref{Segment}.

\begin{figure*}[!htb]
\centering
 \begin{minipage}{.5\textwidth}
  \centering
  \includegraphics[scale=0.53]{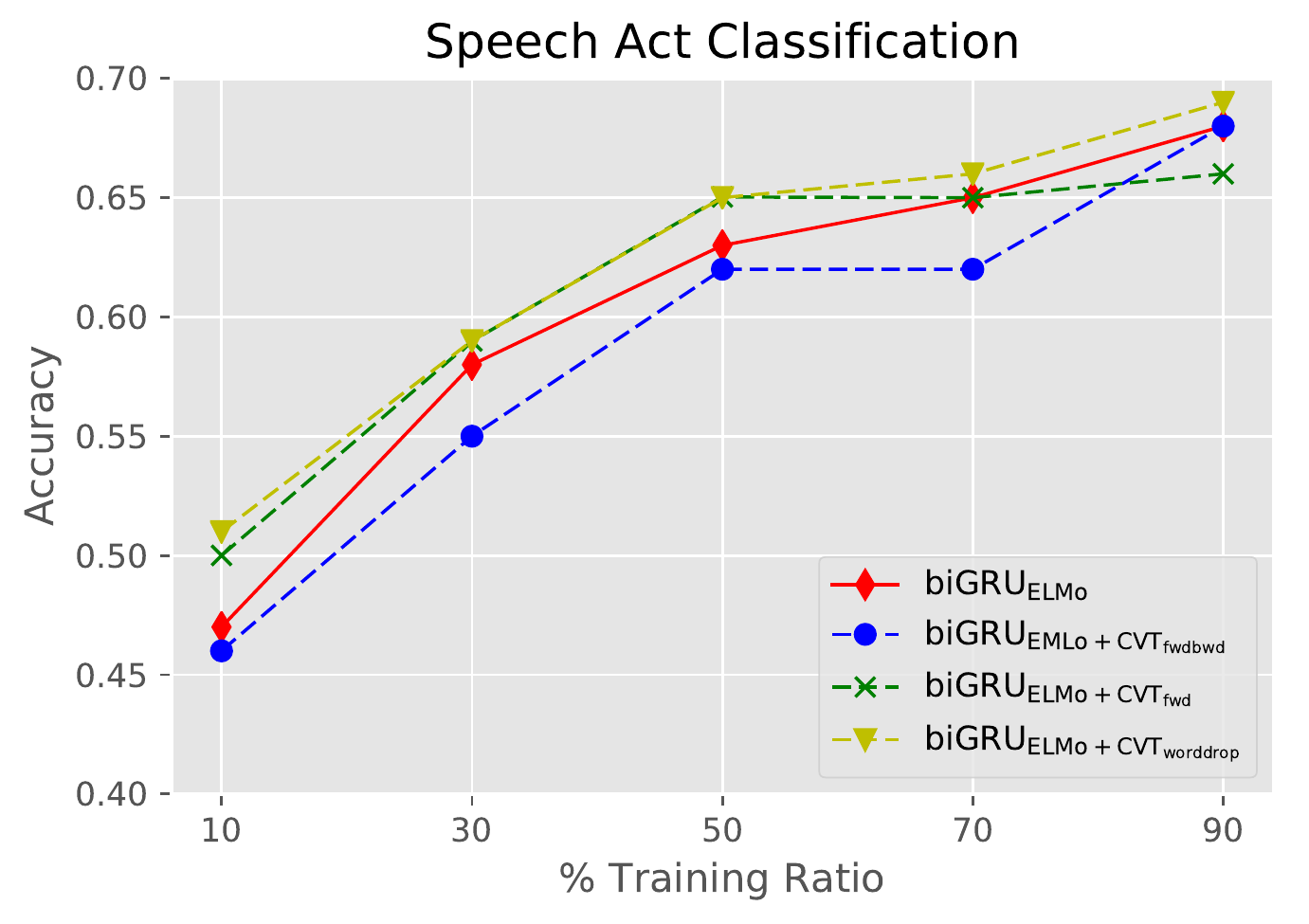}
\end{minipage}%
\begin{minipage}{.5\textwidth}
  \centering
  \includegraphics[scale=0.53]{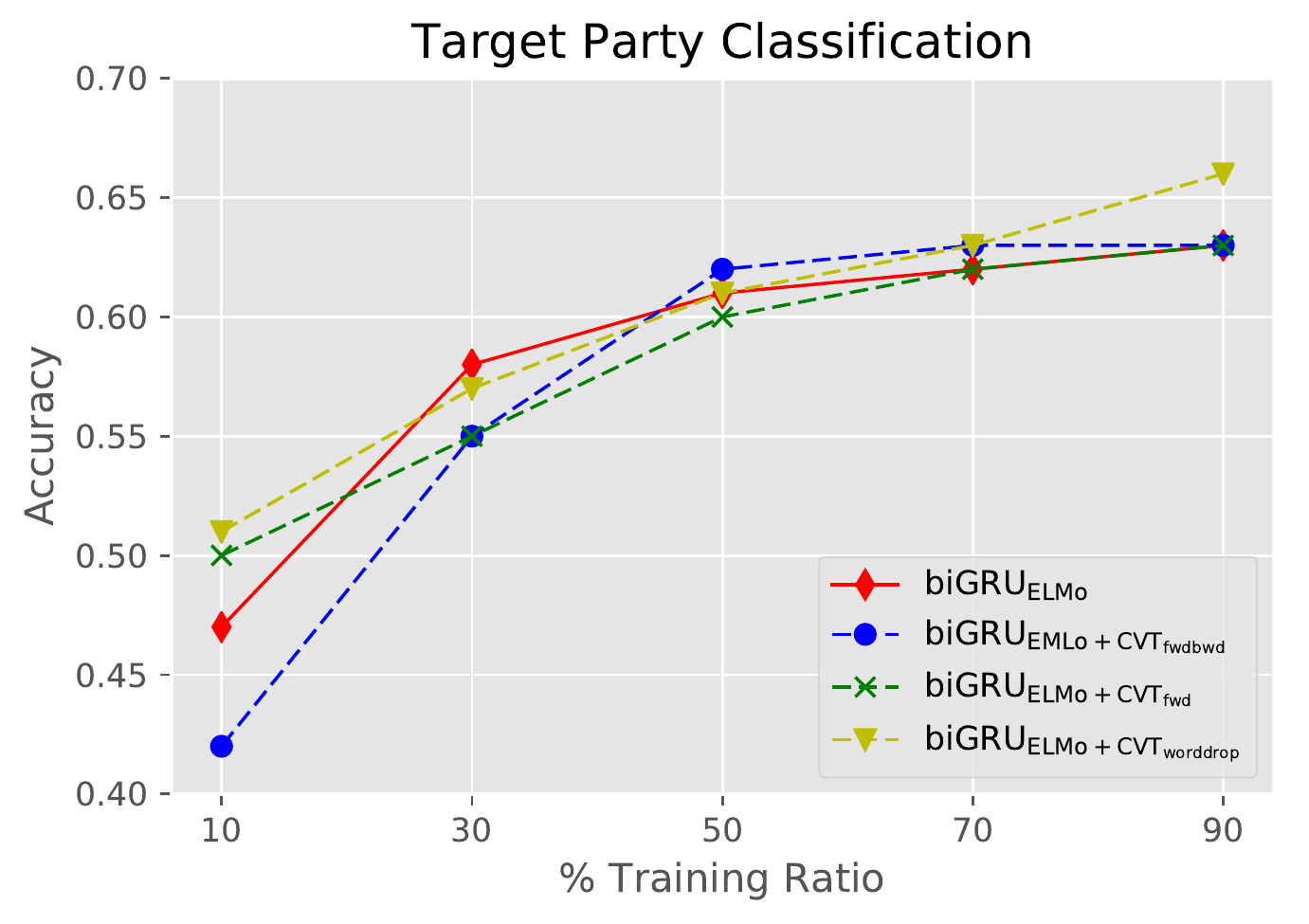}
  \end{minipage}
   \begin{minipage}{.5\textwidth}
  \centering
  \includegraphics[scale=0.53]{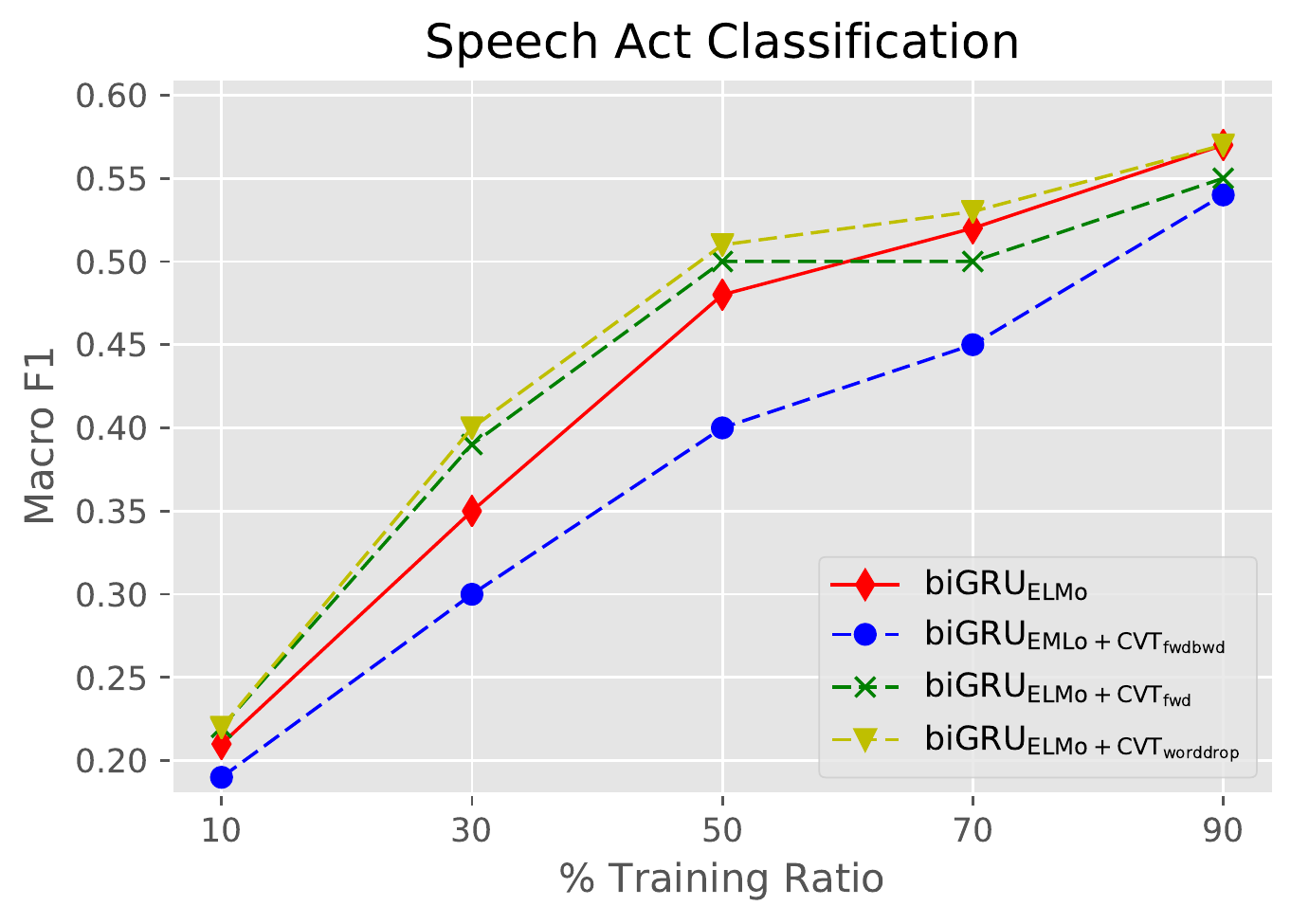}
\end{minipage}%
\begin{minipage}{.5\textwidth}
  \centering
  \includegraphics[scale=0.53]{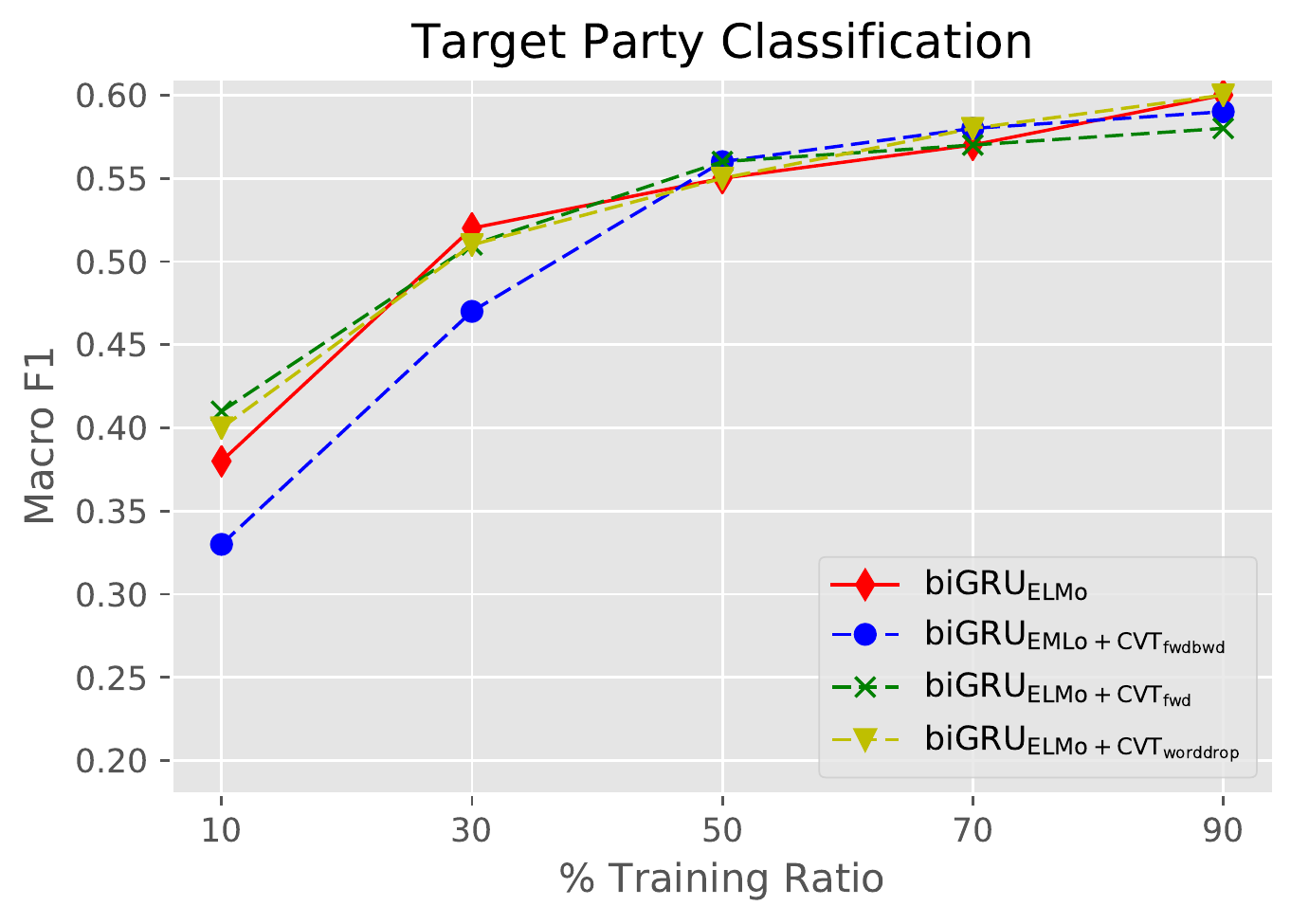}
  \end{minipage}
\captionof{figure}{Classification performance across different training ratios. Note that 90\% is using all the training data, as 10\% is used for validation.}
\label{ssl:results}
\end{figure*}

From the results in \tabref{tab:res}, we observe that the biGRU\footnote{The biGRU model uses ReLU activations, a 128d hidden layer for speech act classification and 64d hidden layer for target party classification, and dropout rate of 0.1.} performs better than the other approaches, and that ELMo contextual
embeddings (biGRU$_{\textbf{\smaller{ELMo}}}$) boosts the performance
appreciably. Apart from ELMo, the semi-supervised learning methods (biGRU$_{\textbf{\smaller{ELMo + CVT$_\text{worddrop}$}}}$) provide a boost in performance for the target party task (wrt accuracy) using all the training data. biGRU$_{\textbf{\smaller{ELMo + CVT$_\text{worddrop}$}}}$ and biGRU$_{\textbf{\smaller{ELMo + CVT$_\text{fwd}$}}}$ provide gains in performance for the speech act task, especially with fewer training examples ($\leq$ 50\% of training data, see \figref{ssl:results}). Performance of semi-supervised learning models with cross-view training (which leverages in-domain unlabeled text) is compared against biGRU$_{\textbf{\smaller{ELMo}}}$, which is a supervised approach. Results across different training ratio settings are given in \figref{ssl:results}. From this, we can see that biGRU$_{\textbf{\smaller\text{ELMo + CVT$_\text{worddrop}$}}}$ and biGRU$_{\textbf{\smaller\text{ELMo + CVT$_\text{fwd}$}}}$ performs better than biGRU$_{\textbf{\smaller\text{ELMo + CVT$_\text{fwdbwd}$}}}$ in almost all cases. With a training ratio $\leq 50\%$, biGRU$_{\textbf{\smaller\text{ELMo + CVT$_\text{worddrop}$}}}$ achieves a comparable performance to biGRU$_{\textbf{\smaller{ELMo + CVT$_\text{fwd}$}}}$.

Multi-task learning
(biGRU$_{\textbf{\smaller\text{ELMo + CVT$_\text{worddrop}$ + Multi}}}$) provides only small improvements for the speech act task. Further, when we add speaker party meta-data (biGRU$_{\textbf{\smaller{ELMo + CVT$_\text{worddrop}$ + Meta}}}$), it provides large gains in performance for the target party task. Overall, the proposed approach (biGRU$_{\textbf{\smaller{ELMo + CVT$_\text{worddrop}$ + Meta}}}$) provides the best performance for the target party task. Its performance is better than the biGRU$_{\textbf{\smaller{ELMo + Meta}}}$ model, which does not leverage the additional unlabeled text using semi-supervised learning, where it achieves 0.70 accuracy and 0.65 Macro F1. Also, ELMo and semi-supervised methods (biGRU$_{\textbf{\smaller{ELMo + CVT$_\text{worddrop}$}}}$ and biGRU$_{\textbf{\smaller{ELMo + CVT$_\text{fwd}$}}}$) provide significant improvements for the speech act task, especially under sparse supervision scenarios (see Figure \ref{ssl:results}, for training ratio $\leq$ 50\%). 

\begin{table*}[!htb]
\centering
\begin{smaller}
\begin{tabular}{p{12cm}cc}
\toprule
 Utterance & Target party & Speaker\\
 \midrule
Our new Tourism Infrastructure Fund will bring more visitor dollars and more hospitality jobs to Cairns, Townsville and the regions. & \class{Labor} & \class{Labor}\\\\[-0.8em]
Just as he sold out 35,000 owner-drivers in his deal with the TWU to bring back the ``Road Safety Remuneration Tribunal". & \class{Labor} & \class{Liberal} \\\\[-0.8em]
Then in 2022, we will start construction of the first of 12 regionally superior submarines, the single biggest investment in our military history. & \class{Liberal} & \class{Liberal} \\\\[-0.8em]
\bottomrule
\end{tabular}
\end{smaller}
\caption{Scenarios where ``Speaker'' meta-data benefits the target party classification task.}
\label{tab:meta}
\end{table*}

\section{Segmentation Results}
\label{Segment}

In the previous experiments, we used gold-standard utterance data, but next
we experiment with automatic segmentation. We use sentences as input,
based on the NLTK sentence tokenizer \cite{bird2009natural}, and
automatically segment sentences into utterances based on token-level
segmentation, in the form of a \texttt{BI} binary sequence
classification task using a CRF model
\cite{hernault2010sequential}.\footnote{We also experimented with a
  neural CRF model, but found it to be less accurate.} We use the
following set of features for each word: token, word shape
(capitalization, punctuation, digits), Penn POS tags based on SpaCy,
ClearNLP 
dependency labels \cite{choi2012guidelines}, relative position in the sentence, and features for
the adjacent words (based on this same feature representation).  We
compute segmentation accuracy (\texttt{SA}:
\citet{zimmermann2006joint}), which measures the percentage of segments
that are correctly segmented, i.e.\ both the left and right boundary
match the reference boundaries. \texttt{SA} for the CRF model is
0.87. Secondly, to evaluate the effect of segmentation on
classification, we compute joint accuracy (\texttt{JA}). It is similar to \texttt{SA} but also requires correctness of the speech act and
target party.  In cascaded style, \texttt{JA} using the CRF model for
segmentation and biGRU$_{\textbf{\smaller{ELMo +
      CVT$_\text{worddrop}$} + Meta}}$ for speech act and target party
classification is 0.60 and 0.64 respectively. Here, segmentation errors
lead to a small drop in performance.

\begin{figure*}[!htb]
\centering
 \begin{minipage}{.65\textwidth}
  \centering
  \includegraphics[height=0.7\linewidth]{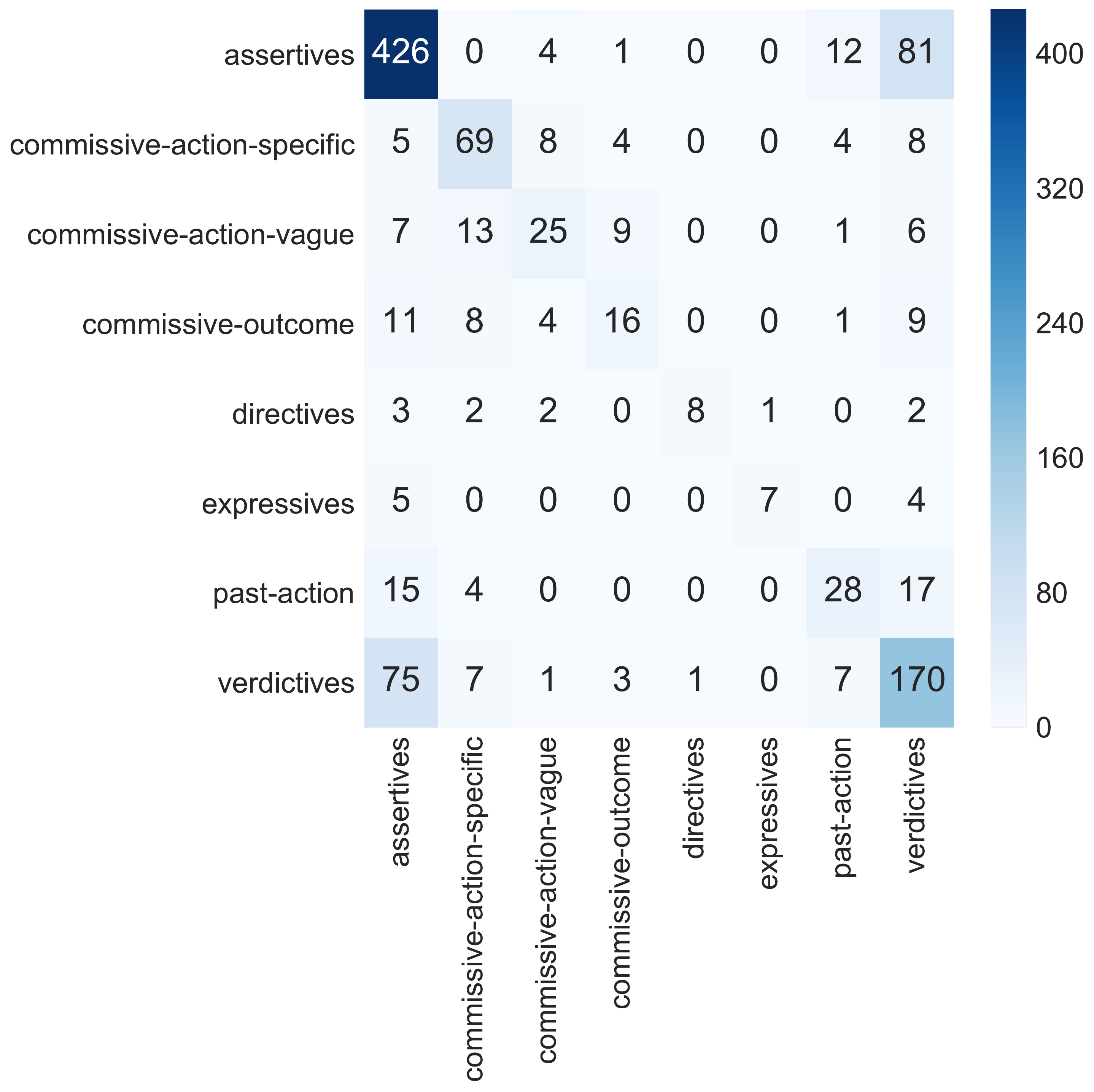}
\end{minipage}%
\begin{minipage}{.35\textwidth}
  \centering
  \includegraphics[height=0.63\linewidth]{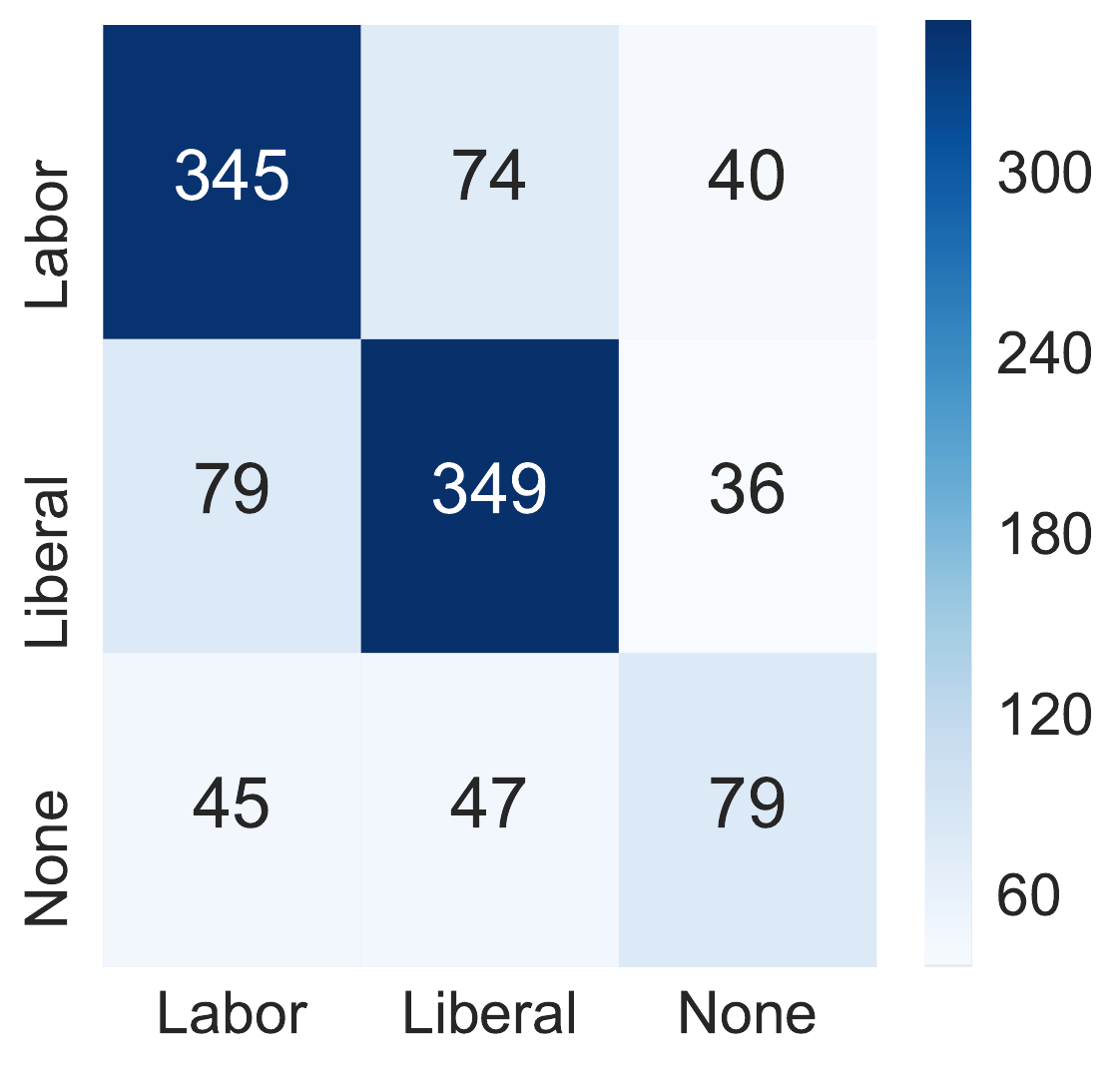}
  \end{minipage}
 \caption{Confusion matrix for speech act and target party classification tasks.}
\label{fig:cm}
\end{figure*}

\section{Error Analysis}
\label{error}

We analyze the class-wise performance and confusion matrix for our best performing approach (biGRU$_{\textbf{\smaller{ELMo + CVT$_\text{worddrop}$} + Meta}}$). Speech act and target party class-wise performance is given in \tabref[s]{tab:dres} and \ref{tab:drest} respectively. We can see that the proposed approach provides improvement across all classes, while achieving comparable performance for \sa{directive}. Recognizing \sa{commissive-outcome} can be seen to be tougher than other classes. In addition, we analyze the results to identify cases where having ``Speaker'' party information is beneficial for predicting the target party of sentences. Some of those scenarios are given in Table \ref{tab:meta}, where the meta-data enables predicting the target party correctly even when there is no explicit reference to the party or leaders.

Confusion matrices for the speech act and target party classification
tasks are given in \figref{fig:cm}. Some observations from the confusion
matrices are: (a) \sa{assertive} and \sa{verdictive} are often
misclassified as each other; (b) \sa{commissive-action-vague} utterances
are often misclassified as \sa{commissive-action-specific}; and  (c)
\class{Labor} and  \class{Liberal} classes are often misclassified as each other for the target party classification task.

\section{Qualitative Analysis}
\label{qa}
\begin{figure*}[!htb]
\centering
 \begin{minipage}{.5\textwidth}
  \centering
  \includegraphics[scale=0.5]{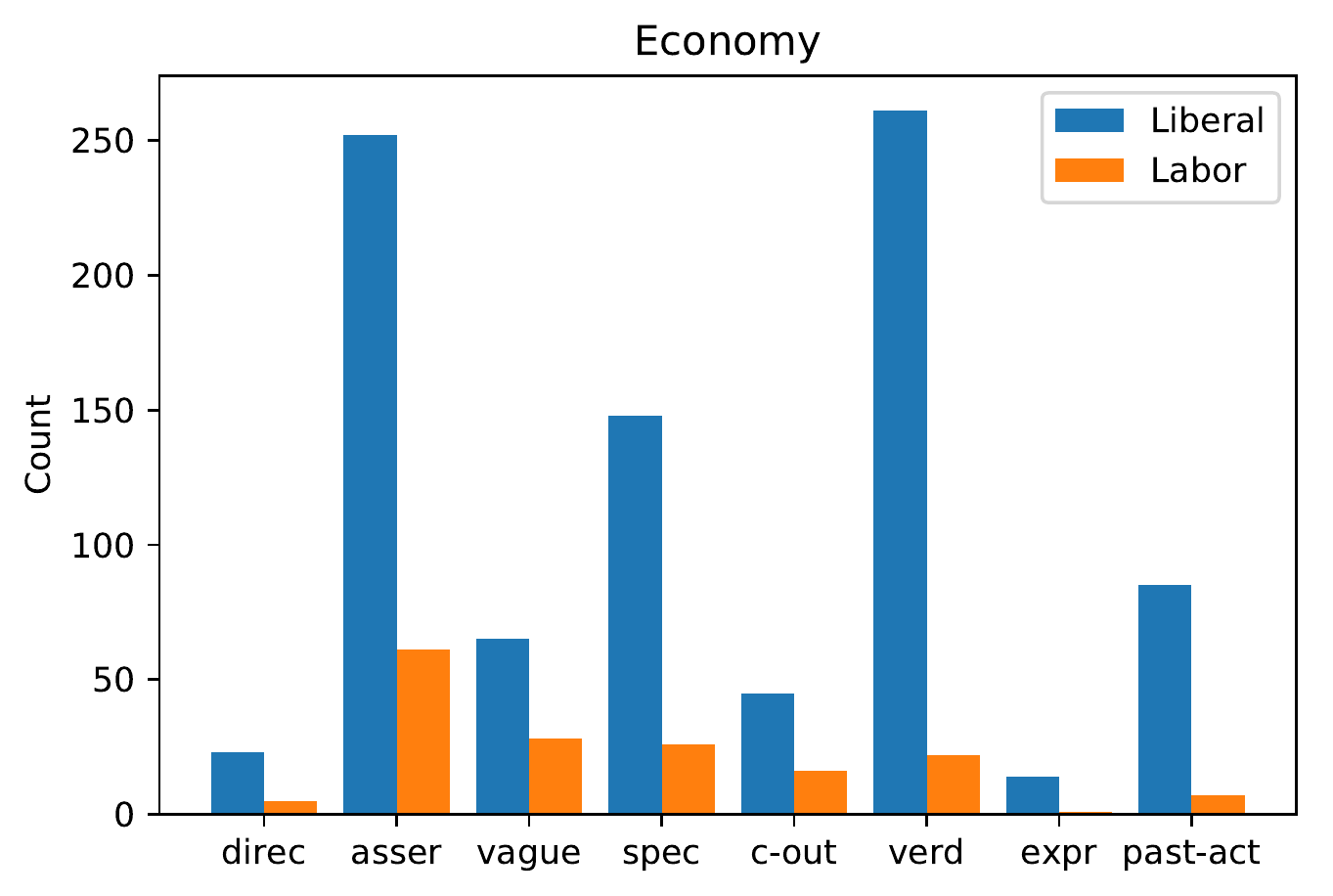}
\end{minipage}%
\begin{minipage}{.5\textwidth}
  \centering
  \includegraphics[scale=0.5]{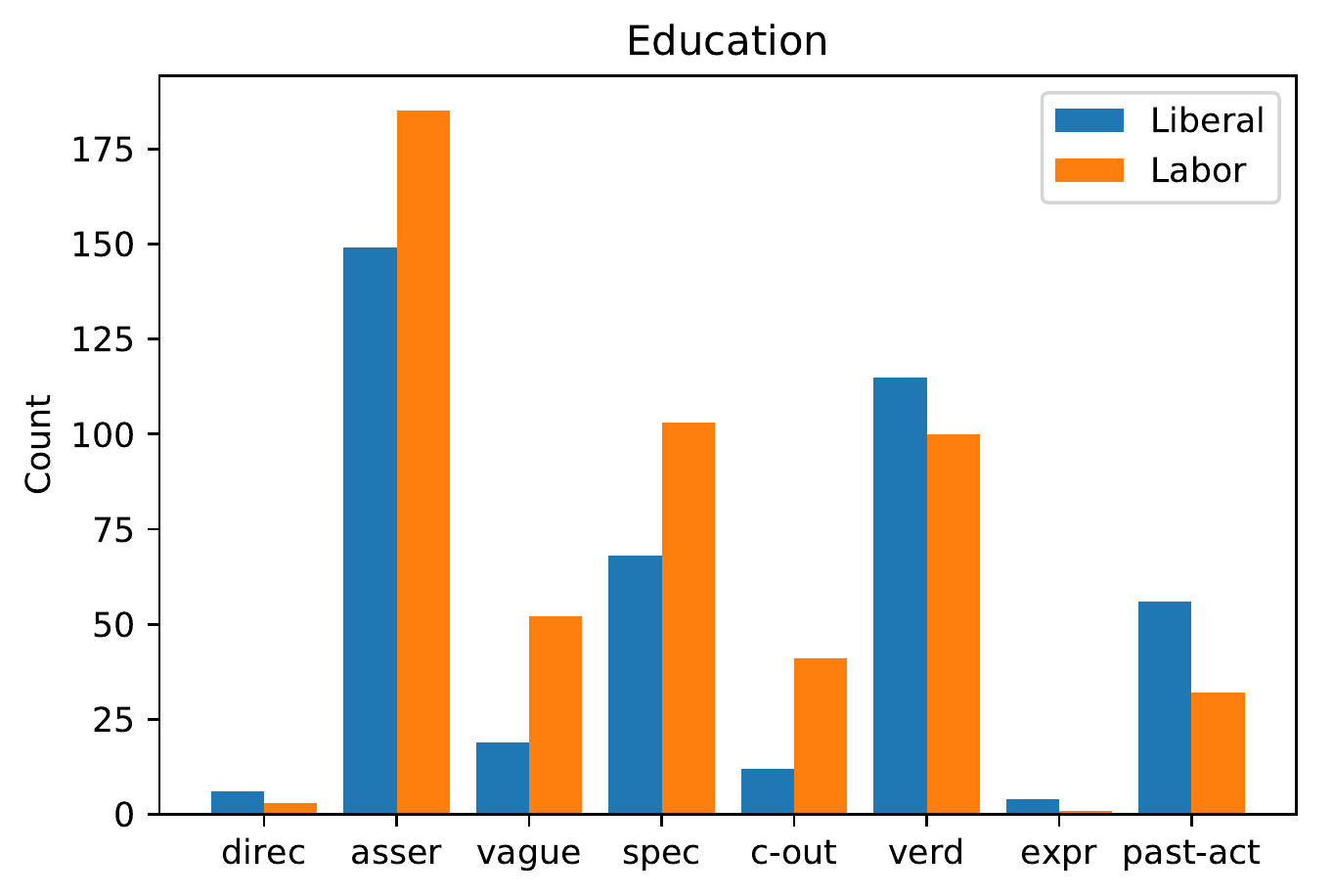}
  \end{minipage}

\begin{minipage}{.5\textwidth}
  \centering
  \includegraphics[scale=0.5]{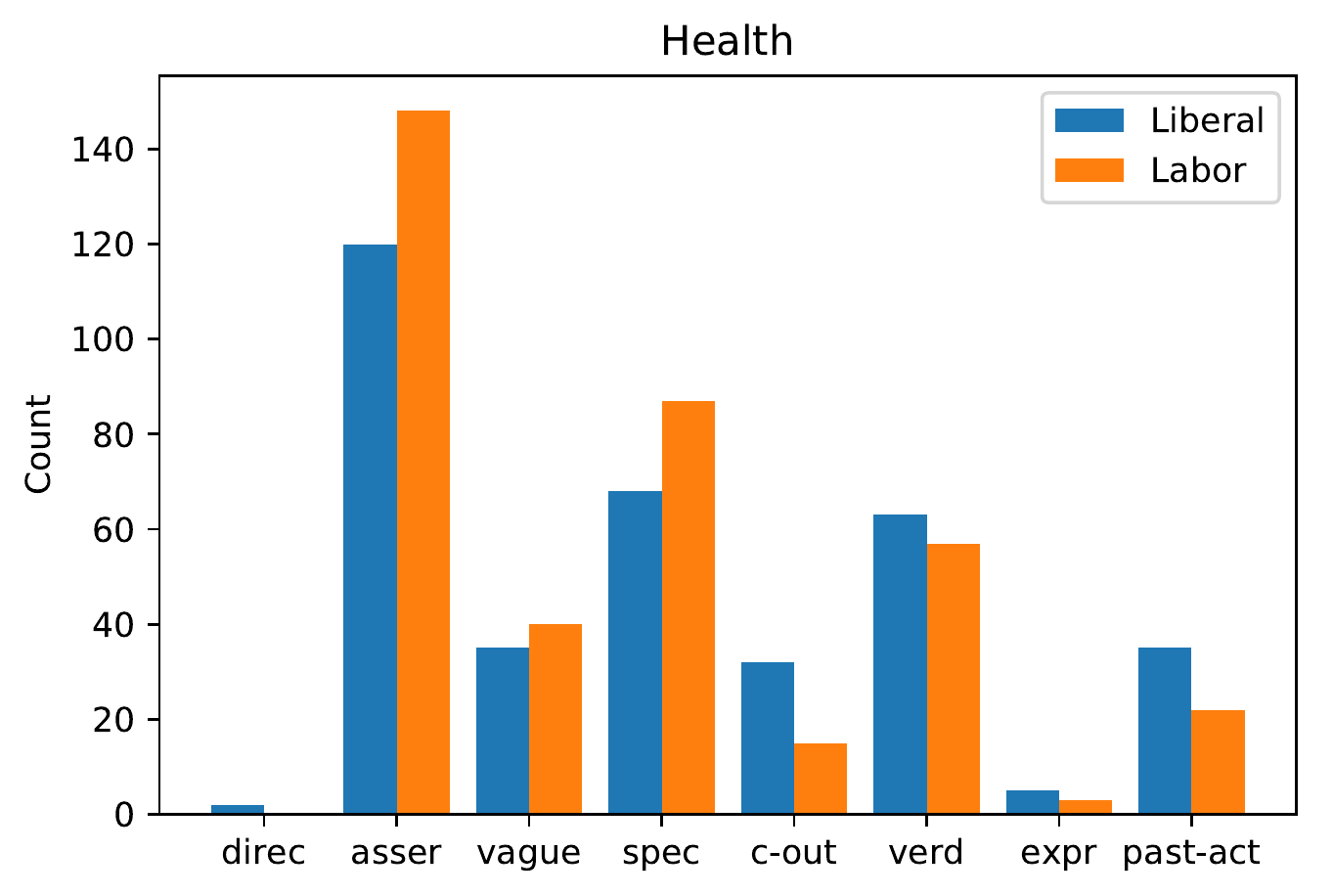}
\end{minipage}%
\begin{minipage}{.5\textwidth}
  \centering
  \includegraphics[scale=0.5]{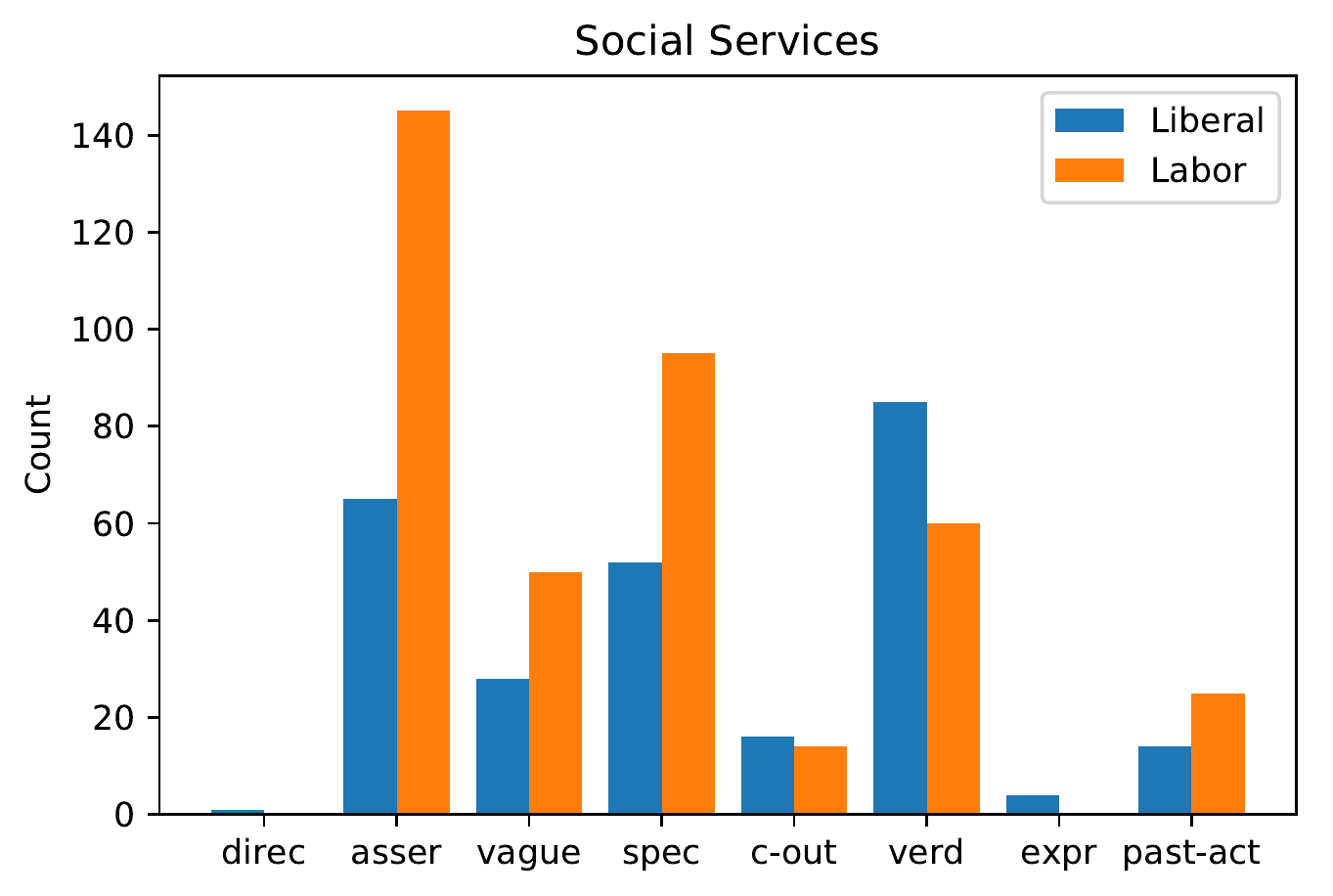}
\end{minipage}

\begin{minipage}{.5\textwidth}
  \centering
  \includegraphics[scale=0.5]{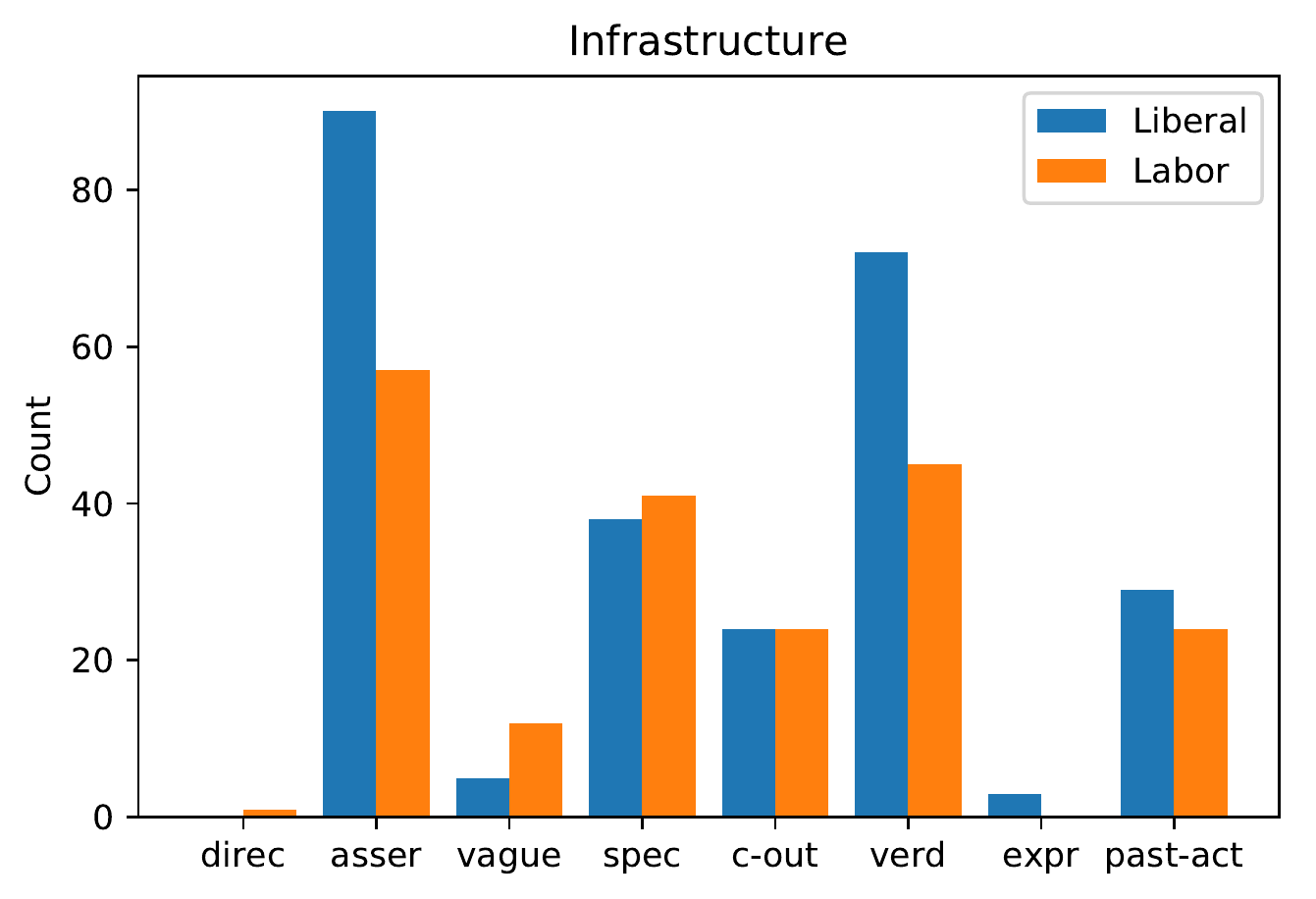}
\end{minipage}%
\begin{minipage}{.5\textwidth}
  \centering
  \includegraphics[scale=0.5]{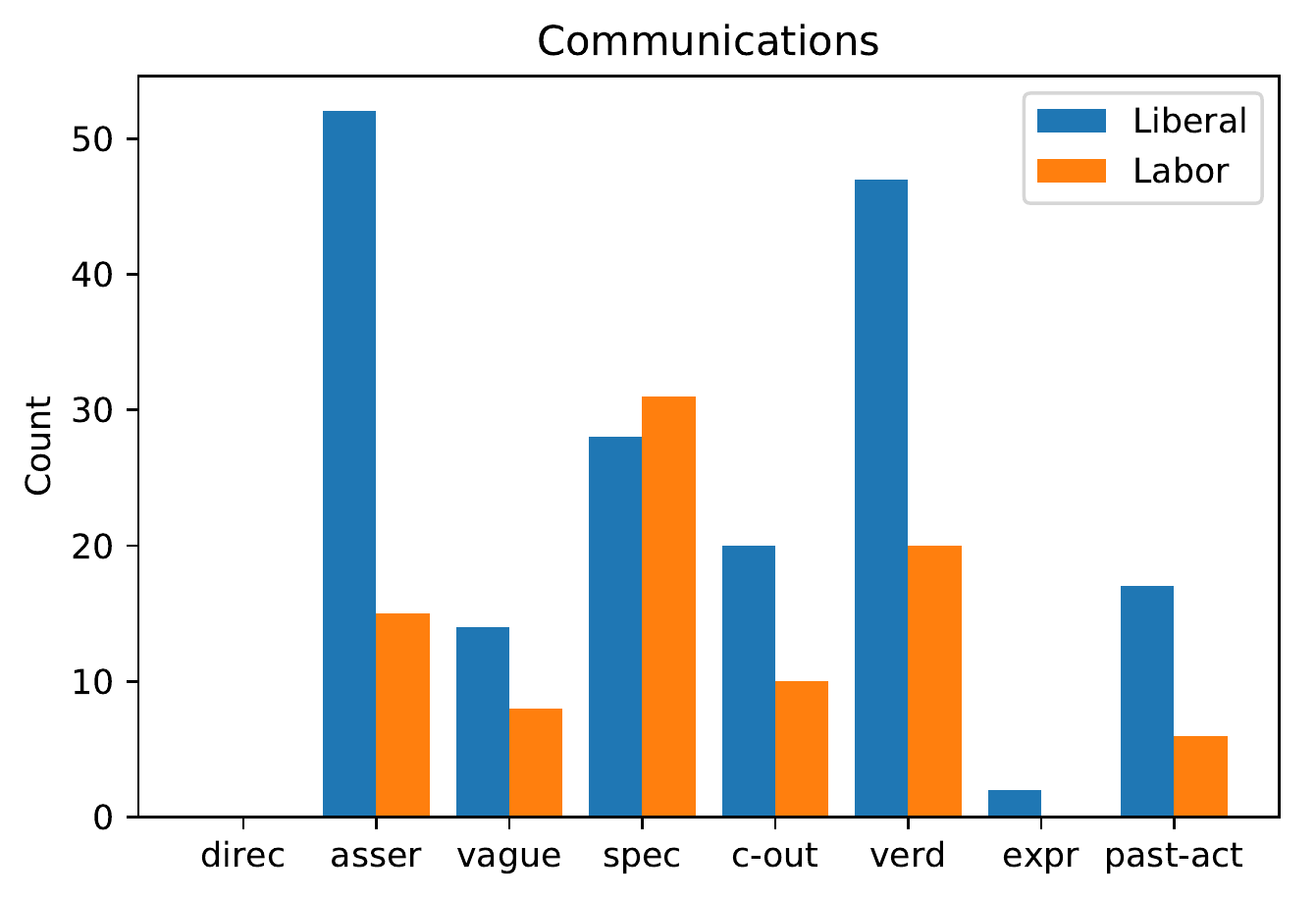}
\end{minipage}
\caption{Policy-wise speech act analysis. Classes include: \sa{directive} (``direc''), \sa{assertive} (``asser''), \sa{commissive-action-vague} (``vague''), \sa{commissive-action-specific} (``spec''), \sa{commissive-outcome} (``c-out''),  \sa{verdictive} (``verd''), \sa{expressive} (``expr''), and \sa{past-action} (``past-act'')}
\label{fig:pol}

\end{figure*}
Here we provide the policy-wise speech act distribution for both
parties, which indicates the difference in their predilection for the
indicated six policy areas (\figref{fig:pol}). We provide results for
the six most frequent policy categories, for each of which, the campaign text is
first classified into one of the policy-areas that are relevant to
Australian politics, by building a Logistic Regression classifier with
data obtained from ABC Fact
Check.\footnote{\url{https://www.abc.net.au/news/factcheck}}
Some observations (based on \figref{fig:pol}) are as follows:
\begin{compactitem}
\item The incumbent government (\class{Liberal}) uses more
  \sa{directive}, \sa{expressive}, \sa{verdictive}, and \sa{past-action} utterances than the opposition (\class{Labor}).
\item \class{Liberal}'s text has
  relatively more pledges (\sa{commissive-action-vague},   \sa{commissive-action-specific} and \sa{commissive-outcome}) on
  \policy{economy} compared to \class{Labor}, whereas \class{Labor} has more pledges on \policy{social
    services} and \policy{education}. This is as expected for right- and left-wing parties respectively. Other policy-areas have a
  comparable number of pledges from both parties. Overall, party-wise
  salience towards these policy areas correlates highly with the
  relative breakdowns in the Comparative Manifesto Project \cite{CMP}:
  where the relative share of sentences from the \class{Labor} and
  \class{Liberal}
  manifestos\footnote{\url{https://manifesto-project.wzb.eu/down/data/2018b/datasets/MPDataset_MPDS2018b.csv}}
  for \policy{welfare state (health and social services)} is 22:7,
  \policy{education} is 9:6, \policy{economy} is 11:23, and
  \policy{technology \& infrastructure (communication, infrastructure)}
  is 17:19.
\item Across policy-areas, \sa{specific} pledges are more frequent than
  \sa{vague} ones. This aligns with previous studies done by
  \citet{Naurin2014} and \citet{APSA2017}.
  
\end{compactitem}

\section{Conclusion and Future Work}
\label{cfw}
In this work we present a new dataset of election campaign texts, based on a class schema of speech acts specific to the political science
domain. We study the associated problems of identifying the referent
political party, and segmentation. We showed that this task is feasible to annotate, and present several models for automating the task.
We use a pre-trained language model and also leverage auxiliary
unlabeled text with semi-supervised learning approach for the target
based speech act classification task. Our results are promising, with
the best method being a semi-supervised biGRU with ELMo embeddings for
the speech act task, and the model additionally incorporating speaker
meta-data for the target party task. We provided qualitative analysis of
speech acts across major policy areas, and in future work aim to expand this analysis
further with fine-grained policies and ideology-related analysis. 

\section*{Acknowledgements}
We thank the anonymous reviewers for their insightful
comments and valuable suggestions. This
work was funded in part by the Australian Government
Research Training Program Scholarship, and
the Australian Research Council. 

\bibliography{naaclhlt2019}
\bibliographystyle{acl_natbib}

\end{document}